\documentclass[letterpaper]{article} 
\usepackage{aaai24}  
\usepackage{times}  
\usepackage{helvet}  
\usepackage{courier}  
\usepackage[hyphens]{url}  
\usepackage{graphicx} 
\usepackage{natbib}  
\usepackage{caption} 
\usepackage{algorithm}
\usepackage{algorithmic}

\usepackage{newfloat}
\usepackage{listings}
\DeclareCaptionStyle{ruled}{labelfont=normalfont,labelsep=colon,strut=off} 
\floatstyle{ruled}
\newfloat{listing}{tb}{lst}{}
\floatname{listing}{Listing}
\usepackage{booktabs}
\usepackage{amsmath}
\usepackage{tcolorbox}
\newtcolorbox{mybox}[2][]{
  title    = {#2},
  colback  = white!5,
  colframe = white!40!black,
  boxsep   = 1mm,
  top      = 1mm, 
  bottom   = 1mm, 
  left     = 1mm,
  right    = 1mm, 
  #1
}

\usepackage[T1]{fontenc}

\title{LLM Voting: Human Choices and AI Collective Decision-Making}
\author{
    Joshua C. Yang,
    Damian Dailisan,
    Marcin Korecki,
    Carina I. Hausladen,
    Dirk Helbing
}
\affiliations{
    Computational Social Science\\
    ETH Zurich\\
    Zürich, Switzerland
}

\begin{document}

\maketitle

\begin{abstract}
This paper investigates the voting behaviors of Large Language Models (LLMs), specifically GPT-4 and LLaMA-2, their biases, and how they align with human voting patterns. Our methodology involved using a dataset from a human voting experiment to establish a baseline for human preferences and conducting a corresponding experiment with LLM agents. We observed that the choice of voting methods and the presentation order influenced LLM voting outcomes. We found that varying the persona can reduce some of these biases and enhance alignment with human choices. While the Chain-of-Thought approach did not improve prediction accuracy, it has potential for AI explainability in the voting process. We also identified a trade-off between preference diversity and alignment accuracy in LLMs, influenced by different temperature settings. Our findings indicate that LLMs may lead to less diverse collective outcomes and biased assumptions when used in voting scenarios, emphasizing the need for cautious integration of LLMs into democratic processes.
\end{abstract}

\section{Introduction}
\begin{figure*}[ht!]
    \centering
    \includegraphics[width=0.7\linewidth]{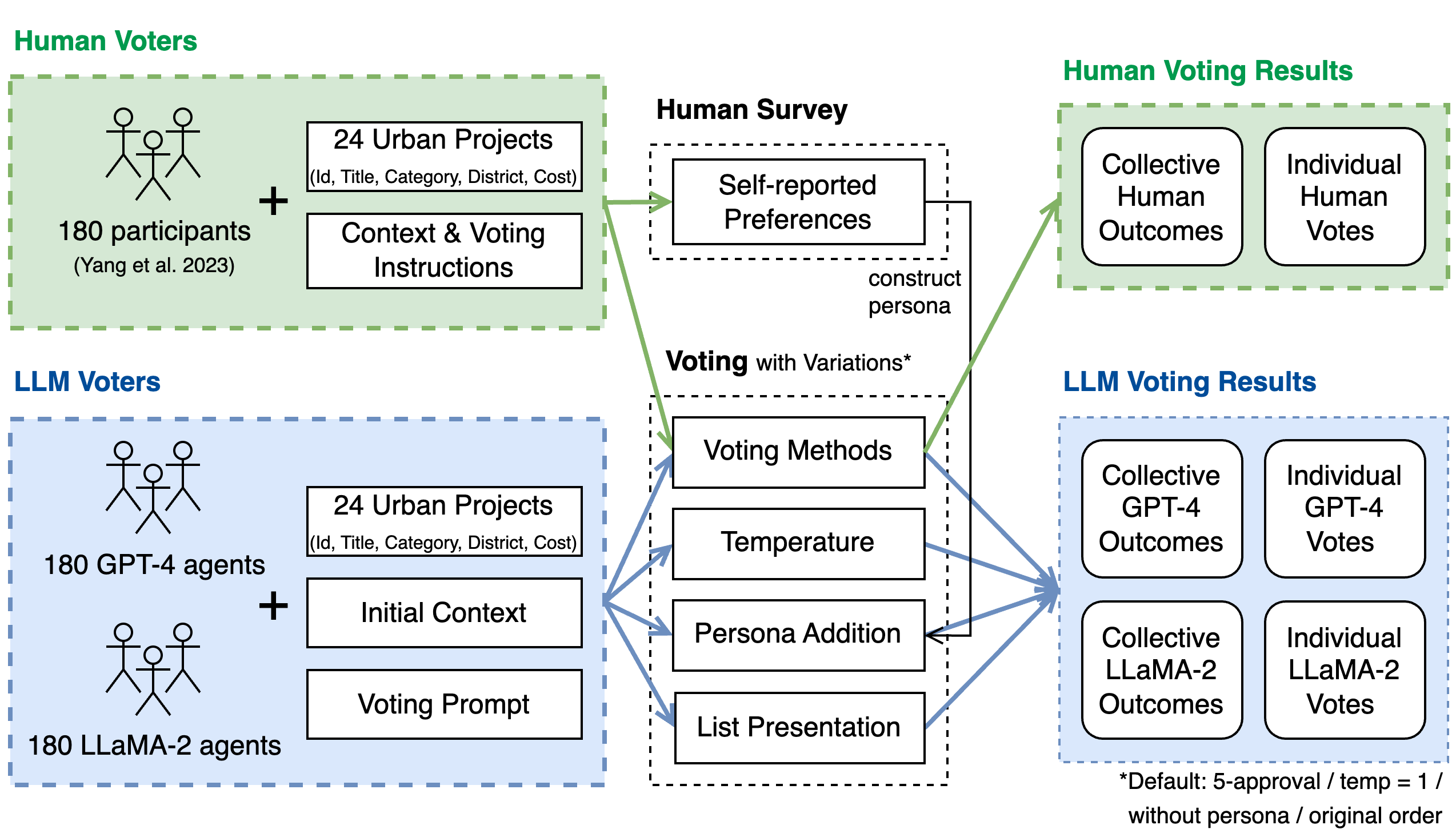}
    \caption{Overview of the LLM voting experimental setup}
    \label{fig:overview-flowchart}
\end{figure*}

Recent breakthroughs in generative models have marked a significant achievement in Artificial Intelligence (AI): the creation of machines capable of fluently processing human language. Large Language Models (LLMs) are increasingly being integrated into a variety of services. As with many discoveries before, a great deal of enthusiasm has been associated with LLMs and their potential applications. As exciting and potentially useful LLMs may be, however, they are also limited in various ways, and their naïve integration may bring many unforeseen consequences. 

LLMs have been used to reproduce classic economic, psycho-linguistic, and social psychology experiments with humans~\citep{aher2023using, argyle_out_2023, horton_large_2023}, as well as behavior on social media platforms~\citep{tornberg_simulating_2023}. Some propose using LLMs to improve online democratic conversations~\citep{argyle_leveraging_2023}. Another proposal is to use LLMs to perform human-related scientific research~\citep{boiko_emergent_2023, bran2023chemcrow, sourati2023accelerating}.

Despite the impressive development of LLMs, however, it is clear that LLMs are currently still limited and prone to errors \cite{sobieszek_playing_2022, floridi2023ai}. The ``hallucinations''~\citep{yao2023llm}, proclivity for lies~\citep{azaria2023internal}, effects such as a strong primacy effect~\citep{wang_primacy_2023}, and clear political biases~\citep{feng_pretraining_2023} all question that LLMs can be safely used in social settings. 

In the area of digital democracy, an application of AI that has sparked considerable debate is the concept of assisted real-time voting. The proposal to use AI ``digital twins'' to replace politicians has been raised~\citep{hidalgo2018}. These AI systems, including customized LLM agents, envisioned to vote and mirror individual voter preferences~\citep{GersbachMartinelli2023}. Proponents believe that such agents could enable more nuanced and granular voting under voter supervision~\cite{plurality2023}. Yet, ethical concerns about automation, democratic integrity, and agent bias necessitate careful consideration. \citet{AllenWeyl2024} argue that Generative Foundation Models (GFMs), such as GPT-4, present unprecedented challenges for democratic institutions. The deployment of such technology in democratic processes would require a cautious approach, ensuring that the core values of democracy are upheld~\cite{helbing2022democracy}.

In our study, we explore the limitations and potentials of integrating LLMs into collective decision-making processes. We focus on examining LLM behavior in voting scenarios by contrasting the voting patterns of human participants with those of LLaMA-2 and GPT-4 agents. This analysis uses data from a Participatory Budgeting (PB) voting experimental study conducted by \citet{yang2024}. The overall setting of the study is illustrated in Fig~\ref{fig:overview-flowchart}. By investigating the behavior of LLMs in this context, we seek to contribute to the broader discourse on LLM decision-making, value alignment, AI explainability, and ethical use of AI in society.

\section{Related Literature}

\subsection{LLM Autonomous Agents}
The popularity of LLMs sparked a trend wherein researchers used LLMs to model "agents" as entities with several submodules. In their landmark paper, \citet{park_generative_2023} show LLM agents interacting with each other in a sandbox community exhibits human-like behavior. \citet{boiko_emergent_2023} suggests combining multiple LLMs for the autonomous design, planning, and execution of scientific experiments. For a literature survey on creating autonomous LLM agents, we refer the reader to \citet{wang_survey_2023}. 

Research into multi-agent LLM systems, drawing inspiration from social and biological cooperative systems, explores the dynamics of cooperative artificial agents \citep{axelrod1981evolution,shoham2009multiagent,bonabeau1999swarm}.
Studies have investigated various LLM agents in both software and embodied forms for complex tasks, such as text evaluation \citep{chan2023chateval}, image captioning, and text-to-image synthesis \citep{zhuge_mindstorms_2023, talebirad_multi-agent_2023, zhang2023building}. Innovations include proactivity and adaptability enhancement \citep{zhang2023proagent}, standardized prompting mechanisms \citep{hong2023metagpt}, roleplaying frameworks for cooperation study \citep{li2023camel}, and debating methods to improve answer quality \citep{wei2022chain, hao_chatllm_2023, du_improving_2023}. 
Furthermore, researchers have explored the use of different personas within a single agent to enhance complex task performance \citep{wang_unleashing_2023}. While most studies focus on surpassing benchmarks in complex tasks, some also focus on the intrinsic characteristics of multi-agent LLM systems \citep{serapio-garcia_personality_2023, liang2023encouraging}.

\subsection{LLM Societies}

One particular example of a multi-agent LLM system is the creation of LLM societies. Such systems create environments for studying debates and societal simulations. \citet{liu_training_2023} introduces an open-source platform for simulating artificial human societies, while \citet{hao_chatllm_2023} proposes a ChatLLM network system for more objective and comprehensive decision-making. \citet{zhuge_mindstorms_2023} explores the concept of natural language-based societies of mind, questioning their optimal governance structure. In \citet{wang_unleashing_2023}, the idea of Solo Performance Prompting (SPP) is proposed, transforming a single LLM into a ``cognitive synergist'' through multi-turn self-collaboration with multiple personas. \citet{du_improving_2023} presents a method where multiple language model instances propose, debate, and refine their responses to reach consensus.

\citet{liang2023encouraging} addresses the Degeneration-of-Thought (DoT) problem with a Multi-Agent Debate framework, where agents engage in ``tit for tat'' arguments and a judge oversees the debate to derive a final solution. \citet{talebirad_multi-agent_2023} introduces a collaborative environment, where intelligent agents with distinct attributes and roles communicate. \citet{wu_autogen_2023} discusses ``Autogen'', an open-source library facilitating conversation among multiple LLM agents. Furthermore, \citet{ferland_rational_2015} and \citet{rank_using_2019} explore the use of personas in civic communication and the potential of digital twins to understand voter behavior and address biases in democratic processes.

\subsection{Social Choice and Rationality}

Social welfare functions play a critical role in aggregating individual preferences into a collective social ranking \cite{sen1995rationality}. This aggregation, denoted by \( R \), derived from individual rankings \( \{R_i\} \), focuses on forming a coherent social order while balancing fairness and rational decision-making \cite{Arrow1951}.

Rationality in social choice theory impacts voting behaviors and preference aggregation. It encompasses:

\textbf{Internal Consistency of Choice}: This aspect, fundamental in decision theory and economics, necessitates that an individual's choices are coherent and logically consistent \cite{Arrow1951}. In our research, this aspect of rationality is tested by varying voting methods. Previous studies find that the voting methods impacts voter preferences and behaviors to a varying extent depends on the context \citep{yang2024, hausladen2023legitimacy}.
    
\textbf{Self-Interest Maximization}: Central to classical economic theory is the assumption that individuals act to maximize their self-interest \citep{Becker1976}. 
Additionally to consistency, self-interest is added to the characterization of an agent  \citep{walsh1994}.
The latter principle is typically modeled as a utility function, where individuals choose options that maximize their utility.

\textbf{Maximization in General}: Rationality extends to the maximization of a defined objective function \cite{sen1995rationality}. In the context of voting, this often implies maximizing collective utility or welfare, where an individual selects the option that yields the greatest collective benefit. This is connected to the idea of other-regarding preferences \citep{blanco2011within}, describing a situation where an individual's utility is impacted by the well-being of others.

Our study explores these three dimensions of rational choice, analyzing their influence on individual and collective voting behaviors.

\subsection{Voting in Multi-Winner Systems}
\label{ss:voting-theory}

Collective actions in modern democracies are becoming increasingly individualized and issue-driven, leading to a state of "chaotic pluralism" that challenges traditional democratic processes \cite{Margetts2016HowAction}. Multi-winner systems, wherein voters select multiple candidates from a broad field with the possibility of multiple winners \cite{skowron2019, elkind2017properties}, offer a potential solution by aligning with pluralistic societal values and diverse demands.

Multi-winner systems translates individual voter preferences into a collective decision, often seeking to maximize collective utility or fairness \cite{aziz2017polynomial}. Approval voting is one common method, where the top candidates receiving the most votes are elected \cite{brams1983approval,laslier2016strategic}.

In the context of LLMs, the presentation of voting options as a list is a critical factor. Similar to how the arrangement of candidates on a ballot can influence voter perception, the order and format in which LLMs present multiple options may affect the selection process, reflecting the need to understand how LLMs mimic these complex multi-candidate and multi-winner decisions.
            
\subsection{Opinion Sampling and Persona Creation}

Our methodology, which compares actual population opinions with LLM-generated responses, is in line with the work of \citet{durmus_towards_2023}. They developed the GlobalOpinionQA dataset and a metric for comparing LLM and human responses, which is crucial to assess how well LLMs reflect human opinions. Similarly, \citet{tornberg_simulating_2023} simulated social media environments using digital personas based on demographic data. Their findings suggest that this approach leads to healthier, less divisive online discussions.

\subsection{LLMs and Human Behaviour}

Exploring the intricacies of political biases and personality traits in LLMs is crucial for their ethical application in human interactions. \citet{serapio-garcia_personality_2023} investigate the role of personality traits in LLMs, developing methods to assess and customize them in advanced models. \citet{feng_pretraining_2023} delve into political biases of LLMs, examining their origins from diverse pretraining data and effects on tasks like hate speech detection. Meanwhile, \citet{argyle_out_2023} explore how LLMs can positively influence online political discourse, with AI chat assistants improving conversation quality and receptiveness to different views. In addition, \citet{wang_primacy_2023} examine the primacy effect in ChatGPT, finding that: i) ChatGPT's decisions are sensitive to the order of labels in prompts, and ii) there is a higher tendency for ChatGPT to select labels in earlier positions as the answer. \cite{abdurahman2024perils} also finds that AI models resemble a single WEIRD (Western, Educated, Industrialized, Rich, Democratic) individual rather than simulating a diverse pool of participants. In summary, these studies underscore the significance of recognizing personality traits in LLMs and managing biases to ensure their ethical and effective use. The study by \citet{wei2022chain} demonstrates that ``Chain-of-Thought''(CoT) prompting, which involves providing intermediate reasoning steps, enhances the problem-solving abilities of LLMs. In their experiments, CoT significantly improved performance on arithmetic, commonsense, and symbolic reasoning tasks.

\subsection{LLMs and Collective Intelligence}

As language models (LLMs) are increasingly used to mimic human behavior, recent research has also begun exploring the collective behavior of LLMs. The work of \citet{jarrett2023language} formalizes the problem of digital representation as the simulation of an agent's behavior to yield equivalent outcomes from the collective decision-making mechanism. \citet{chuang2024wisdom} examines whether groups of LLMs can mimic human behavior when role-playing as partisan personas (like Democrats or Republicans). The authors prompted the LLMs to role-play as different personas created with varying levels of background detail. The research finds that incorporating CoT reasoning or a lack of detailed persona tends to diminish the wisdom of partisan crowds effect, making the group less likely to converge to more accurate beliefs.

\citet{fish2023generative} merges social choice theory with the text generation abilities of LLMs. This framework is designed to facilitate complex decision-making, such as selecting textual statements that represent collective preferences. A recent work by \citet{gudino2024large} suggests that LLMs using augmented data can more accurately predict the preferences of an entire participant population compared to probabilistic samples, which may not be representative. They indicates that LLMs have significant potential for constructing systems of augmented democracy.

\section{Methods}

\begin{figure}[ht!]
    \centering
    \includegraphics[width=0.85\linewidth]{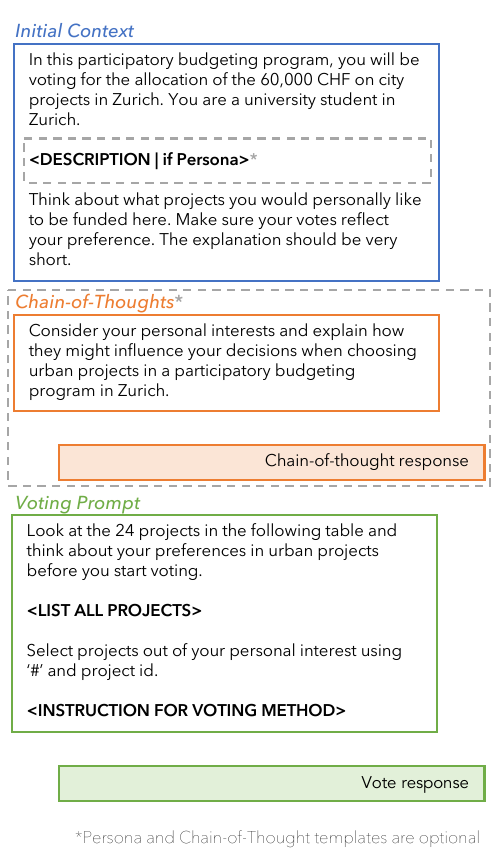}
    \caption{Overview of the LLM prompt template.}
    \label{fig:prompt}
\end{figure}
In this paper, we compare the behavior of human voters in a PB setting to that of LLM voters. For human voters, we used data from \citet{yang2024} who simulated a Participatory Budgeting (PB) scenario in Zurich in an online experiment. The dataset\footnote{https://github.com/joshuay1/zurich-pb-voting} contains 180 university student participants' votes on urban projects, reflecting real urban interests for allocating a CHF\,60,000 budget across 24 projects.
We simulated the voting of human participants using LLaMA-2 and GPT-4 Turbo models. These models take in an \textbf{\emph{Initial Context}} with instructions and background that the human voters also received in their experiements, followed by a \textbf{\emph{Voting Prompt}} for casting votes on the projects in a format that can be processed (see Fig.~\ref{fig:prompt}). 

Both the human participants and LLM agents were presented the same list with 24 projects (Table 3 in the appendix) in the voting process, detailed by ID, Name, District (Nord, Süd, Ost, West), Urban Category (Nature, Culture, Transportation), and Cost (CHF\,5,000 or CHF\,10,000). There is exactly one project for each combination of these characteristics.

\subsubsection{Testing Different LLMs:} Our study uses two types of agents: LLaMA-2 70B (Q8\_0) and GPT-4 Turbo (GPT-4-1106-preview). We chose the open-source LLaMA-2 to ensure other researchers can replicate and verify our experiments, thereby enhancing the reliability of our research. In contrast, GPT-4 Turbo was selected for its widespread use and relevance in current applications, making our findings more applicable to real-world scenarios. 

\subsubsection{Aggregation:} In each experimental setting, we replicated the voting process with 180 LLM voters, matching the sample size of the human experiment. LLM votes, formatted with a `\#' and project ID in the textual responese, were parsed using \texttt{regex}. The Borda Count method was applied to Ranked votes to convert ranks into points, while the 10-point Cumulative votes were normalized to ensure equal weighting. Outcomes were then aggregated, ordering projects by total votes or points for a direct comparison with human voter results.

\subsection{Voting Methods}
\label{sec:vote_instructions}

We explore the four most representative voting input methods in multi-winner settings, as identified by \citet{yang2024} in their voting experiment from which the data originate. Among these, 5-Approval voting is used as our baseline due to its common usage in PB settings and its simplicity, which facilitates the comparison of outcomes when each voter is constrained to select an equal number of projects. For each voting method, we use the same textual voting instructions that were given to human voters for the LLM agents. The voting instructions are as follows:

\begin{itemize}
    \item \textbf{Approval}: \emph{``Select any number of projects. Here, in this vote, you can select all the projects you approve of."}
    \item \textbf{5-Approval}: \emph{``Select exactly 5 projects."}
    \item \textbf{Cumulative}: \emph{``Distribute 10 points among the projects you like. List the projects and the points you allocate, separated by a colon."}
    \item \textbf{Ranked}: \emph{``Select 5 projects and rank them from the most preferred to the 5th most preferred."}
\end{itemize}
Note: The conversion from ranks to points in \textbf{Ranked} voting follows the conventional Borda’s Method. (1st rank: 5 points, 2nd rank: 4 points, and so forth)

\subsection{Evaluation of Voting Outcomes}
In our study, we use three indices to compare voting outcomes from LLM agents and human participants:
\begin{itemize}
    \item \textbf{Aggregated preferences}: We employ Kendall's \(\tau\) to assess the similarity between LLMs and humans: \mbox{\(\tau = \frac{2}{n(n-1)} \sum_{i<j} \text{sgn}(x_i - x_j) \text{sgn}(y_i - y_j)\)}, where \( n \) is the number of observations, and \( x_i, x_j \), \( y_i, y_j \) represent ranked preferences.
    \item \textbf{Individual vote comparison}: We utilize the Jaccard Similarity index, defined as \(J_i(A, B) = \frac{|A \cap B|}{|A \cup B|}\), where \( A \) and \( B \) are the vote sets from the i\textsuperscript{th} LLM and human agents, respectively.
    \item \textbf{Preference diversity}: The complement of Jaccard Similarity (i.e. Jaccard Distance), \(D_{J_i}(A, B) = 1 - J_i(A, B)\), quantifies the diversity of preferences of voters. A higher value indicates greater heterogeneity in voting patterns.
\end{itemize}

\subsection{Experiments}

\subsubsection{Variation of the Temperature Parameter:} The temperature parameter \(t\) modulates the probability of choosing the next token following \mbox{\(p(\mathrm{token}_i)=\frac{\exp{\mathrm{logit_i}/t}}{\sum_j \exp{\mathrm{logit_j}/t}}\)}. We systematically vary the temperature parameter within the range \(t\in[0,2]\) with 0.5 increments, noting that the standard temperature setting in ChatGPT is 1. This helps assess the impact of different randomness levels on the model's outputs.
    
\subsubsection{Variation of List Presentation:} Our study focuses on two list presentation variations in the context of multi-candidate elections:
\begin{enumerate}
    \item \textbf{Primacy Effect:} This investigates how the positioning of items at the beginning of a list affects their choices.
    \item \textbf{Numerical Labeling Effect:} This investigates the impact of numerical IDs on the LLM's decision-making process.
\end{enumerate}

By examining both the primacy effect and the numerical labeling effect, the study aims to disentangle the influences of item ordering and numerical labeling on LLM decision-making.

\subsubsection{Addition of Persona} We construct persona descriptions from participant responses in the survey data conducted in the same study of human votes. From the survey data, we have the self-reported rating on urban category preferences (Culture, Nature, Transport), district (Nord, Süd, Ost, West), and importance of decision-making factors (district, urban category, cost). These details are incorporated into the \textit{Initial Context} to introduce a variation of persona. By relying on self-reported preferences rather than personal demographic data, we effectively simulate diverse voter profiles for the assisted-voting scenario, thus avoiding the ethical complexities and privacy concerns associated with AI stereotyping and the use of sensitive personal information.

Each entry consists of the participant's ID, urban category preferences, district information, connection to the city, and importance ratings. The process of persona generation involves processing survey data and using the numerical scores to generate descriptive human language labels that reflect the participants' preference intensity. The function further combines this with participants' district preferences and their prioritization of decision factors like district importance, urban category significance, and cost, creating a comprehensive and personalized narrative for each participant.
Here is the persona constructed based on the example in Appendix C:
\begin{quote}\scriptsize\sffamily
You are a university student from Nord district in Zurich.
In urban topics, you have a strong preference 
for transport.
When deciding on projects, you find the district to be moderately important, the urban category very important, and the cost of the project not important at all.
\end{quote}

\subsubsection{Addition of Chain-of-Thought (CoT):} In the attempt to improve performance without further complicating the comparison between the results of human voters and LLM agents, we applied CoT reasoning technique to the best-performing model. In this experiment, the model used was focused on GPT-4. The agents would be prompted an additional time before the voting prompt with the project list (see Fig.~\ref{fig:prompt}).

We specifically prompted agents to reflect on their preferences before presenting the PB projects. This approach aimed to mitigate the issue of ``unfaithful'' CoT explanations, as described by \citet{turpin2024language}. By withholding project information during the initial CoT phase, we prevented the LLM agents from using pre-existing choices to rationalize their decisions. 

For reproducibility, the code and data used in our experiments are available in this GitHub repository: \url{github.com/ethz-coss/LLM_voting}. For technical appendices and the \textbf{appendix} referenced in this paper, see the extended version of this work on arXiv: \url{arxiv.org/abs/2402.01766}.

\section{Results}
\begin{figure}[tbp] 
    \centering
    \includegraphics[width=0.9\linewidth]{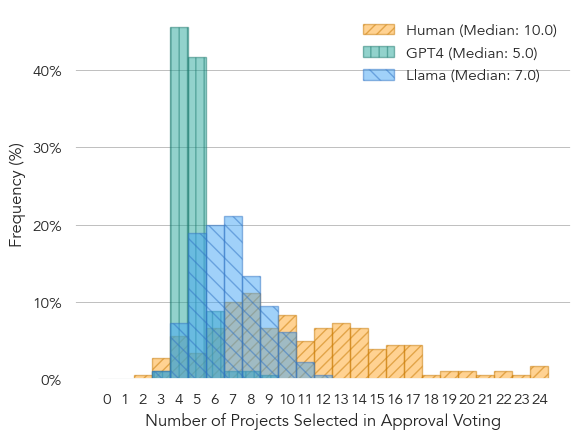}
    \caption{
    Histograms showing the frequency percentage of a certain number of selected projects out of 24 projects in Approval Voting between human and LLM voters.}
\label{fig:appr_hist}
\end{figure}
\subsection{Number of Selected Projects}
The three voter types show distinct behaviors with respect to approval and cumulative voting. As shown in figure~\ref{fig:appr_hist},
human participants display a broad spectrum of {\em approval} patterns, ranging from a minimum of 2 to a maximum of all 24 projects. In contrast, LLaMA's approval patterns form bell-shaped distribution, peaking around 7 projects. GPT4 demonstrates an even narrower distribution, with a median of 5 projects.
In {\em cumulative voting}, LLaMA's point allocations frequently surpassed the prescribed maximum of 10 points, reflecting a deficiency in the model's numerical reasoning capabilities. In contrast, GPT4 adhered strictly to the voting instruction of assigning 10 points.

\begin{figure*}[tb]
\centering
\includegraphics[width=0.9\linewidth]{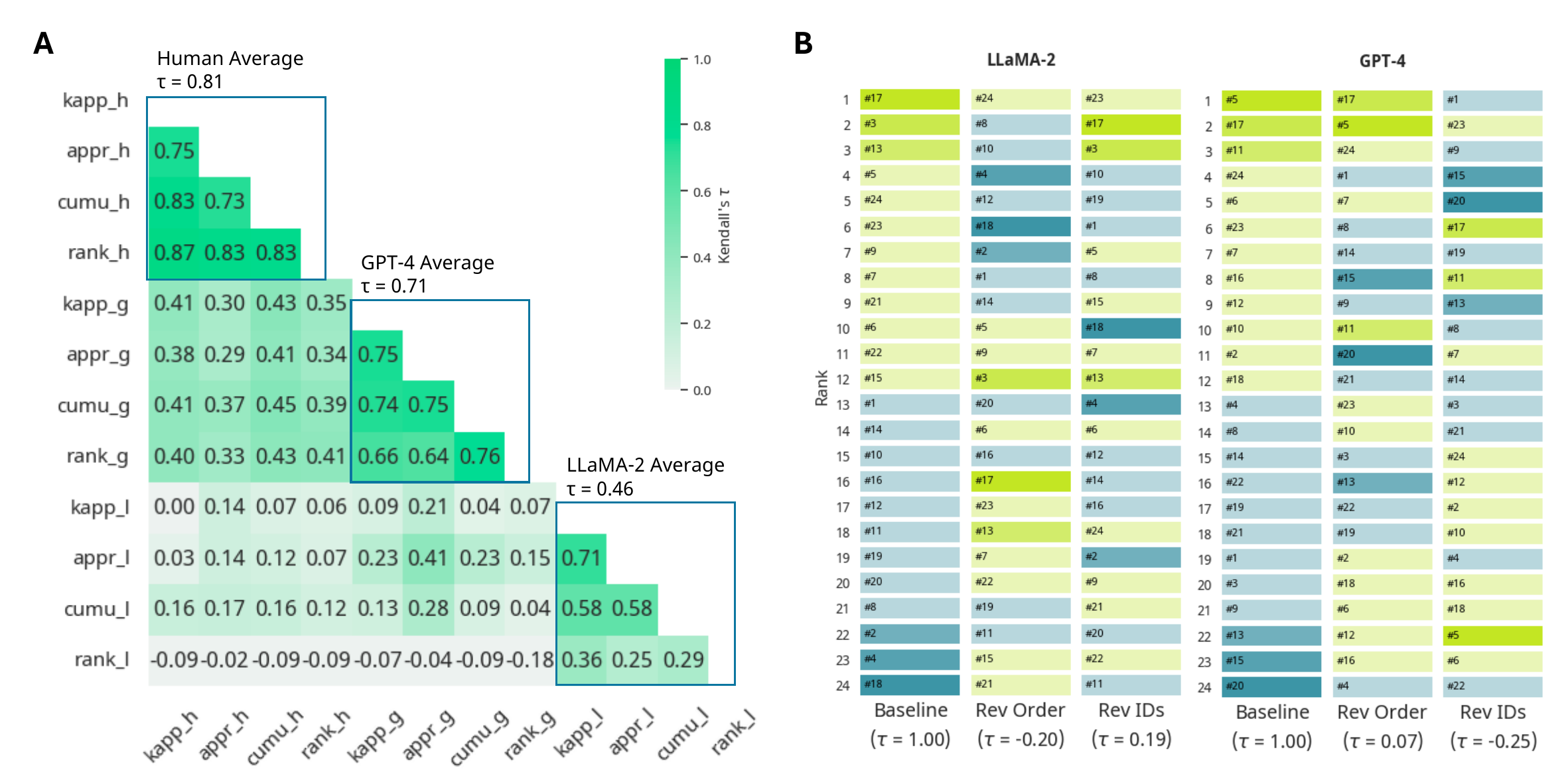}
\caption{
\textbf{A:} Heatmap of Kendall's $\tau$ across different voting methods with votes cast by human (``\_h''), GPT-4 (``\_g''), and LLaMA-2 (``\_l'') voters, showing the similarity between ranking orders of 24 projects, across different groups and methods. Voting methods include \textit{5-Approval} (kapp), \textit{Approval} (appr), \textit{Cumulative (10 points)} (cumu), and \textit{Ranked} (rank).
Higher values of $\tau$ (green) show greater agreement between the compared voting methods.
\textbf{B:} Comparative analysis of the primacy effect on LLM ranking behaviors in 5-Approval voting. 
The outcomes of human votes (baseline) are juxtaposed against results with reversed presentation order or ID sequence, showing how the presentation sequence affects ranking.
Project rankings are color-coded to reflect their relative positioning, with green representing projects in the top half, and blue in the bottom half. In addition, the top and bottom three projects are shown with darker hues. Disparities in $\tau$ values underscore the variable susceptibility of each LLM to ordering effects in vote aggregation.
}
\label{fig:heatmap_all}
\end{figure*}

\subsection{Consistency across voting methods}

Figure~\ref{fig:heatmap_all}A shows the voting consistency of humans and LLMs across voting methods measured by Kendall's $\tau$. This metric ranges from $-1$ for complete disagreement, $0$ indicating no discernible pattern in voting, to $1$ for perfect agreement across the various voting methods.

Human voters displayed high average consistency ($\tau$:~0.81) across different voting input methods, that is, human voters are clear with their preference and vote consistently when different voting input methods are presented. Particularly, ranked voting (rank\_h) exhibited the highest consistency values for human voters, suggesting that the outcome of ranked voting is more representative of the collective human preferences.

GPT-4 also showed considerable consistency ($\tau$:~0.71) across voting input methods, with its preferences most distinctly aligned in cumulative voting (cumu\_g). LLaMA-2 had a lower average consistency score as compared to humans and GPT-4 agents ($\tau$:~0.45). The LLaMA-2 $\tau$ value in ranked voting (rank\_l) was notably lower, indicating the least alignment with other voting methods in their selection of projects 

\subsection{Consistency across list order variations}

To investigate how sensitive the LLMs are to the order of projects, we presented the 24 projects in varying sequences when prompting the LLM: original order, reversed order, and original order with reversed project IDs. Upon examining the responses individually, each agent provided a convincing rationale behind their votes, as denoted in example responses in the appendix. However, the aggregated votes reveal a real shift in the collective outcome that is not obvious in individual votes. 

LLaMA-2 agents demonstrated considerable volatility in preferences due to changing order; the reversed order caused a substantial reshuffling ($\tau=-0.2$). When the IDs were reversed, LLaMA-2 agents with reversed IDs displayed more stability by keeping two of the top three projects (dark green) at the top (Fig.~\ref{fig:heatmap_all}B). GPT-4 agents showed a stronger consistency when the order is reversed, but the reversed ID scenario resulted in a notable disturbance, elevating the lowest-ranked projects (dark blue). Kendall's $\tau$ coefficients at the bottom of Fig.~\ref{fig:heatmap_all}B show that both LLMs experienced a significant drop with reversed orders and IDs.

These results highlight the sensitivity of LLMs to list presentation, aligning with findings where ChatGPT showed a similar sensitivity to label order \cite{wang_primacy_2023}. However, our study did not observe the expected \textit{Primacy Effect}, where top-listed items are typically preferred. As shown in Table 1 in the appendix, in our voting scenarios, there was no significant correlation between project rankings and their IDs, which represented their order on the list. These findings also emphasize the importance of considering the aggregated outcome to understand the potential impact of LLM agents in democratic settings, beyond individual voting behaviors and LLM-generated rationales.

\subsection{Incorporating Persona into LLM Voting}

The personas of the LLM agents are constructed based on human survey responses regarding the importance of urban categories and residential districts as well as the prioritization of district characteristics, urban categories, and decision-related costs. 

When personas are applied via a static prompt to the LLM models, individual votes become more aligned to human votes for both, LLaMA-2 (from $J=0.14$ to $0.21$) and GPT-4 ($J=0.18$ to $0.30$). 

The incorporation of personas not only influences individual sets of votes but also increases the similarity in collective project rankings.
Notably, GPT-4 with persona achieves the highest $\tau$ of 0.54. Moreover, GPT-4 voters without persona adjustments ($\tau=0.39$) still outperform LLaMA-2 voters with persona variations ($\tau=0.12$). LLaMA-2 without persona creates an outcome ($\tau=-0.04$) that is akin to a random outcome.
Qualitatively, GPT-4 without persona exhibits a pronounced bias towards transportation projects. With persona adjustments, however, GPT-4's voting distribution across various project types becomes more diverse and therefore more aligned with human outcomes (Fig.~\ref{fig:human_diff}B).

\subsection{Incorporating Chain-of-Thought into LLM Voting}

\begin{figure*}[tb]
    \centering
    \includegraphics[width=\linewidth]{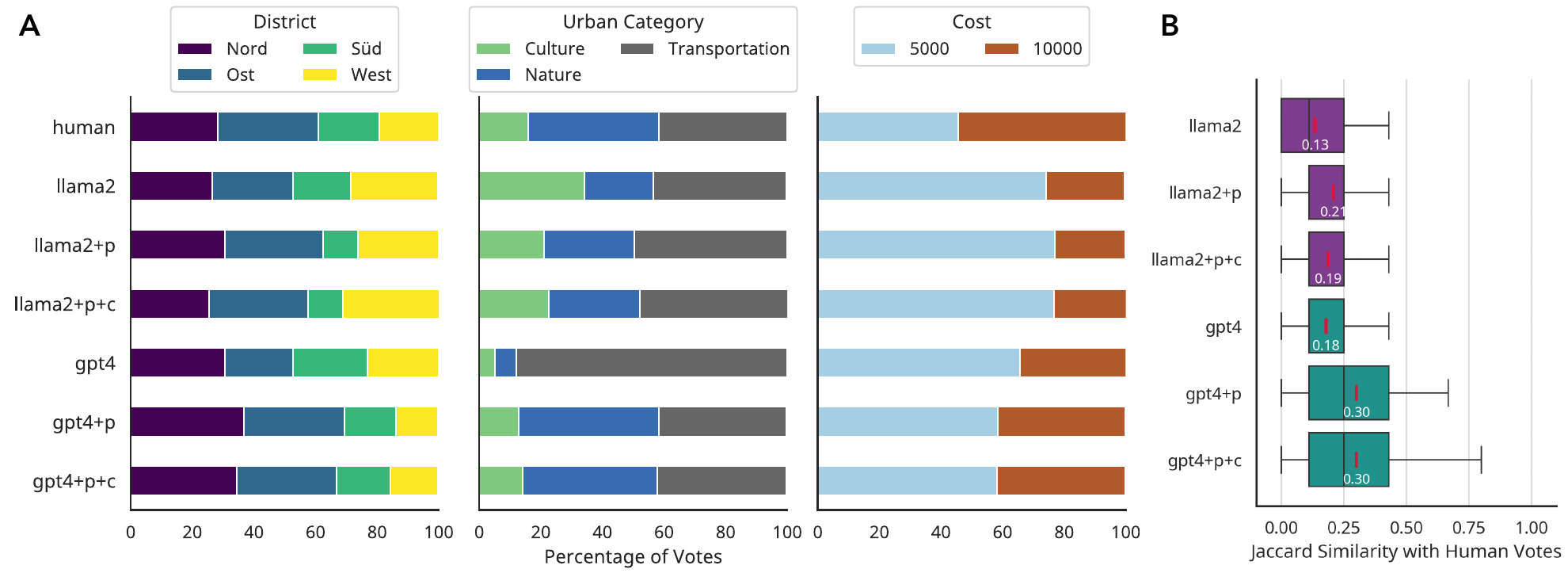}
    \caption{\textbf{A:} Stacked bar plots displaying the distribution of 5-approval votes among different districts (left) and urban project categories (right) for various voter types, including human participants, LLaMA-2, GPT-4, and their respective enhancements (persona, CoT). The addition of persona and CoT is denoted as ``+p'' and ``+c'' respectively. Each bar's segment denotes the proportion of votes contributed by each voter type to the district, category, or cost. \textbf{B:} Box plot of Jaccard Similarity of \textit{Individual} human votes against the agent votes. The addition of persona and CoT is denoted as ``+p'' and ``+c'' respectively. The short red lines and annotated numbers denote  mean values. The closer agents' votes are to human votes, the closer the value is to 1.0.}
    \label{fig:human_diff}
\end{figure*}

We analyze whether the inclusion of CoT reasoning enhances collective decision-making in LLMs. Our findings indicate that CoT has minimal impact on both individual and collective decision-making levels. As illustrated in Figure~\ref{fig:human_diff}A, GPT-4 and LLaMA-2 agents that underwent the CoT process demonstrate similar collective preferences to those without CoT prompting. Figure~\ref{fig:human_diff}B reveals that for GPT-4, Jaccard Similarity values remain consistent with or without CoT. For LLaMA-2, there’s even a slight decrease in similarity (0.21 to 0.19) when CoT is used. The statistical analysis reinforces these observations, showing no significant differences for either mode (GPT-4: $T\text{-statistic} = 0.0119$, $P\text{-value} = 0.9906$; LLaMA-2: $T\text{-statistic} = 1.6506$, $P\text{-value} = 0.0997$).

However, in analyzing the responses qualitatively, we find that the addition of CoT has great potential to improve the explanability of AI decision-making processes. 

For example, GPT-4 agents in 5-approval voting scenarios return simple selections of projects, such as: 

\begin{quote}\scriptsize\sffamily
\#19,\#20,\#21,\#22,\#7
\end{quote}

This output fails to clarify the rationale behind the decisions. By contrast, incorporating CoT provides a more comprehensive thought response. For example, here is the thought response of agent 1 before voting (Full response in appendix):

\begin{quote}\scriptsize\sffamily
As a university student from the Nord district in Zurich with a strong preference for transport within urban topics, my decisions in a participatory budgeting program would be heavily influenced by proposals tackling issues of mobility and connectivity. 
Hence, I would favor projects that aim to improve the transportation infrastructure, such as the development of bike lanes, enhancement of public transport services, and the integration of sustainable transport solutions. 
Projects that specifically target improvements within the Nord district would be of slightly less importance to me than the overall category of urban transport, but I would still consider them favorably, especially if they promise to better the daily commute and accessibility for residents, including my fellow students.
The cost of the project would not influence my decision, as I am focused on the potential benefits and improvements in the transport sector rather than financial constraints. My main criterion for voting for a project would be its potential to make travel within Zurich more efficient, safer, and environmentally friendly.
\end{quote}

This additional thought response offers valuable insights into the reasoning behind the decisions for AI explainability. The implications of this effect are further discussed in the subsequent discussion section.

\subsection{Qualitative Vote Comparison to Human Voters}

There are some notable differences in the proportion of votes allocated to certain types of projects by the LLaMA-2, GPT-4, and human voters (Fig.~\ref{fig:human_diff}A).
LLaMA-2 agents demonstrated a tendency to favor more budget-conscious options, shown in the cost bar plot in Fig.~\ref{fig:human_diff}A, often selecting projects costing around CHF\,5,000. Human voters are the least cost-conscious voter in this study, with the highest votes favoring the more expensive CHF\,10,000 projects.

GPT-4 with no persona variation, in contrast, exhibits a strong preference in urban category for transportation-related projects. The intensity of this preference towards transportation is not shared with human voters. This discrepancy may stem from an over-reliance on the default ``university student'' demographic profile, coupled with a stereotypical assumption about the preferences of this group. Incorporating persona variations into GPT-4 agents helps mitigate this bias, leading to voting outcomes that more closely aligned with those of human voters.

For a more detailed comparison, see Figure 8 in the appendix, which lists project-specific differences. The voting patterns of GPT-4 agents align slightly more with human preferences than those of LLaMA-2 agents, with average deviations of 101.9\% and 141.9\%, respectively. LLaMA-2 agents exhibit a distinct preference for kid-oriented projects (\#3 and \#9), diverging from human choices. Meanwhile, GPT-4 agents without persona variation heavily favor projects enhancing bicycle infrastructure (\#5, \#7, \#11).

\begin{figure}[h!]
    \centering
    \includegraphics[width=0.9\linewidth]{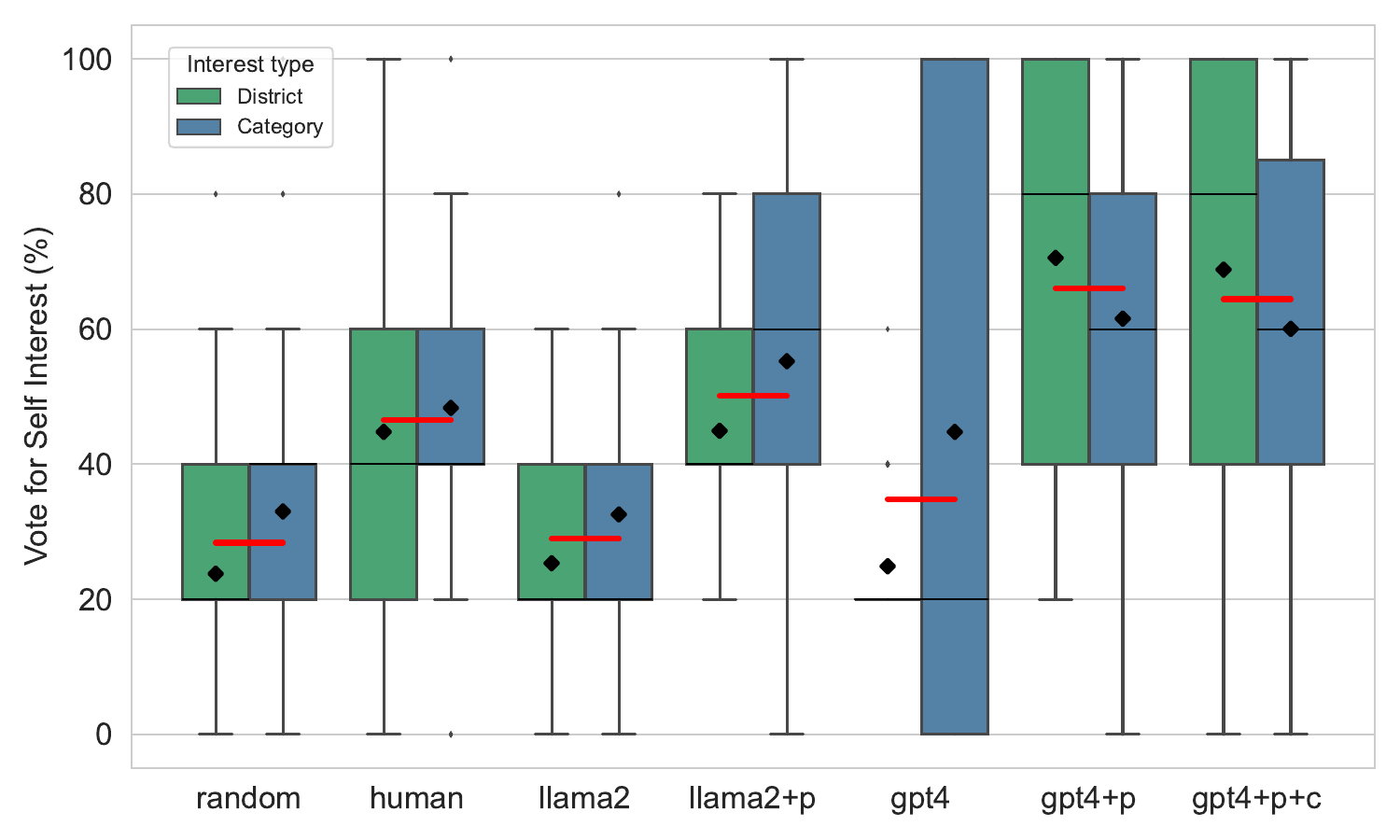}
    \caption{Box plots illustrating the distribution of vote percentages towards the voters' self-identified district and category interests. The red lines represent the median values, while the diamond symbols denote the mean values for each group. A random voting situation is added for a clearer comparison with voting outcomes that consist of a particular preference.
    }
    \label{fig:self}
\end{figure}

\subsection{Alignment between Stated Preference and Votes}

We also analyzed how votes align with surveyed preferences of human voters. As shown in Figure~\ref{fig:self}, human voters typically allocate nearly half of their votes to projects within their self-reported district and preferred urban category, indicating openness to supporting projects outside their primary interests.

In contrast, LLaMA-2 agents without persona variation exhibit voting patterns akin to random outcomes. For both LLaMA-2 and GPT-4 agents, integrating personas leads to a shift toward choices more aligned with stated preferences and human votes. However, when personas are incorporated, GPT-4 agents tend to over-align with self-reported districts and urban interests, surpassing the human tendency for self-prioritization. This overfitting suggests an excessive adherence to the characteristics of the provided human preferences.

\subsection{Diversity in Different Temperature Settings}

\begin{figure*}[tb]
    \centering
    \includegraphics[width=0.9\linewidth]{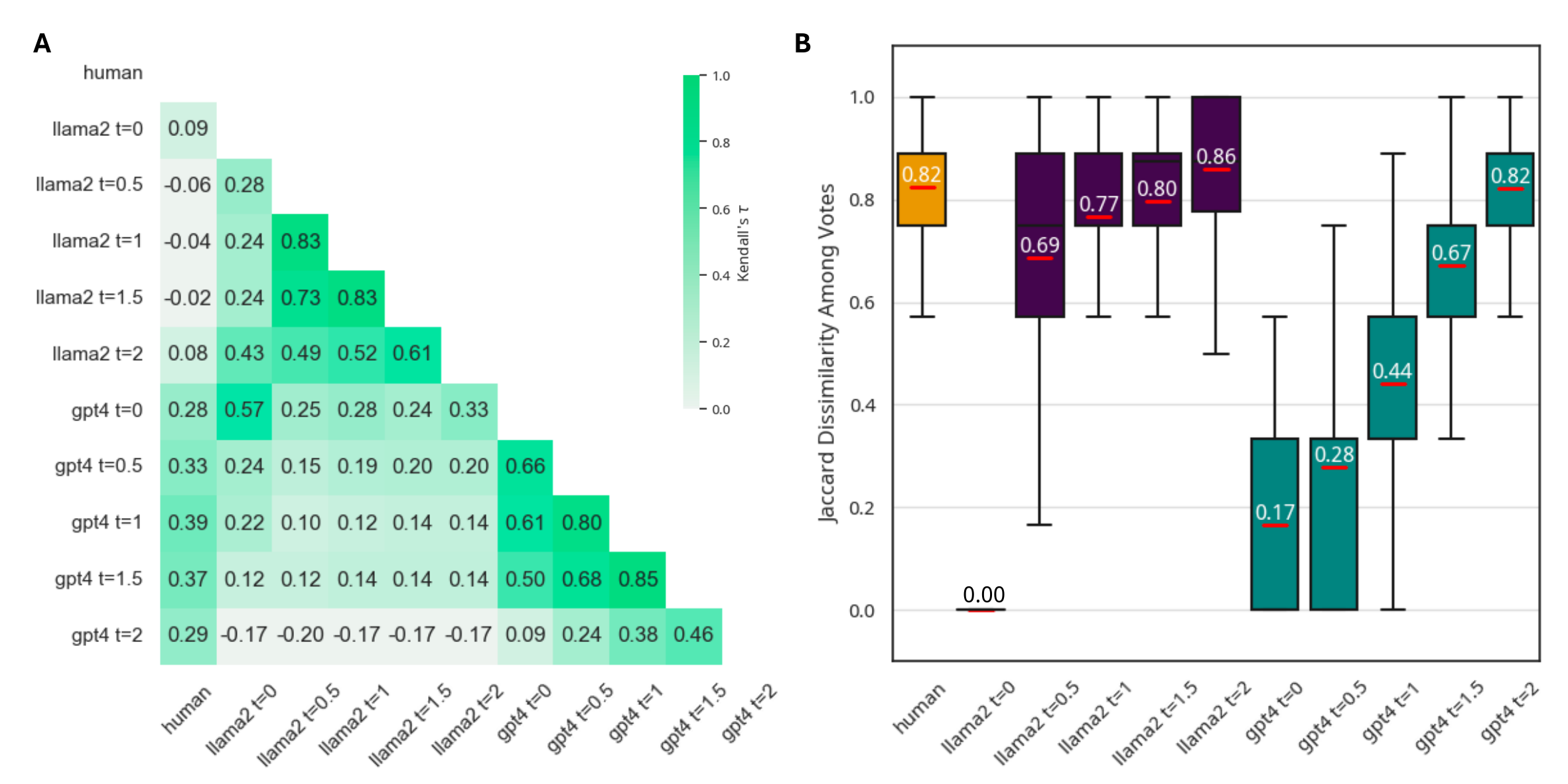}
    \caption{\textbf{A:} Heatmap displaying Kendall's $\tau$ coefficient across various temperature settings of LLaMA-2 and GPT-4 models, compared directly with human vote outcomes. Each cell represents the similarity between the \textit{collective} ranked results of voting sets, with temperature denoted by $t$. Values closer to 1 indicate a higher similarity in the overall ordering of outcomes, whereas values closer to 0 indicate randomness.    
    \textbf{B:} Box plot illustrating the Jaccard Dissimilarity between all \textit{individual} votes from human voters and various temperature settings of LLaMA-2 and GPT-4 models. The mean for each group is indicated by a short red line, with the numerical value displayed. Dissimilarity values closer to 1 suggest greater average deviation between vote sets, indicating less similarity in voting patterns.}
    \label{fig:temp}
\end{figure*}

Analysis of LLM agents' voting behaviors across different temperature settings reveals distinct trends in how these settings influence collective decision-making processes. At lower temperatures, both GPT-4 and LLaMA-2 agents demonstrate highly concentrated preferences, consistently selecting similar project sets. This behavior aligns with the deterministic nature of lower temperatures, where the LLMs favor the most probable outcomes based on their training, leading to limited diversity in choices.

As temperatures increase, there is a notable shift towards broader and more diverse preferences. This is particularly evident at a temperature of 2, where LLM agents begin to explore a wider array of options, reflecting a range of human interests (Fig.~\ref{fig:temp}B). The increased randomness allows the models to simulate a more human-like variance in preferences, although this comes at the cost of reduced predictability in outcomes.

For our analysis, we simplified our description by equating the limit as temperature approaches zero ($t\to+0$) with a temperature of 0. Mathematically, as $t\to+0$, the softmax function converges to argmax, selecting the token in a perfectly deterministic manner. This behavior was observed in our experiments with LLAMA-2 but notably not with GPT-4 in Figure~\ref{fig:temp}B, suggesting a possible divergence in their softmax implementation.

The alignment of LLaMA-2 and GPT-4 agents across different temperature settings with human voting patterns is illustrated in Figure~\ref{fig:temp}A. Among the LLaMA-2 and GPT-4 voters themselves, outcomes generated with the temperature setting of 1.0 are found to be the most stable, as they consistently exhibit high Kendall's $\tau$ values above 0.8 compared to settings of 0.5 and 1.5. 

When we compare LLM outcomes with human outcome, however, the Kendall's $\tau$ values are relatively low across all temperature settings for LLaMA-2, indicating a consistent lack of alignment with human voting outcomes. GPT-4 agents achieve a better alignment with human outcome at a temperature setting of 1.0, as evidenced by the highest $\tau$ values, 0.39, in the first column of Figure~\ref{fig:temp}A.

As expected, increasing the temperature setting in GPT-4 introduces greater diversity in the voting outcomes but also leads to more pronounced deviations from human-like decision-making. For instance, when the temperature is raised from 1.0 to 1.5, there is only a minor reduction in Kendall's $\tau$ from 0.39 to 0.37, which suggests a slight decrease in alignment with human outcomes. At the same time, this increase in temperature causes a significant rise in the Jaccard Dissimilarity of the votes, from 0.44 to 0.67, as shown in Figure~\ref{fig:temp}B. In the subsequent discussion, we will assess whether the modest reduction in alignment is an acceptable compromise for the substantial gain in response diversity.

The alignment of LLaMA-2 and GPT-4 agents across different temperature settings with human voting patterns is illustrated in Figure~\ref{fig:temp}A. Among the LLaMA-2 and GPT-4 voters, outcomes generated with the temperature setting of 1.0 seems to be the most stable, as they are relatively consistent with that of 0.5 or 1.5 with $\tau$ all above 0.8. 

For LLaMA-2 agents, the Kendall's $\tau$ values are close to 0 across all temperature settings, indicating a consistent lack of alignment with human voting outcomes. In contrast, GPT-4 agents achieve optimal alignment at a temperature setting of 1, as evidenced by the highest $\tau$ values. Overall, when a higher temperature introduces more diversity into the voting outcomes, it also results in a greater deviation from human-like decision-making patterns. However, increasing the temperature to 1.5 results in only a slight decrease in $\tau$ from 0.39 to 0.37, yet affects the Jaccard Dissimilarity or diversity of the votes, as shown in Figure~\ref{fig:temp}B, jumps from 0.44 to 0.67. The temperature of 1.5 seems to balance the need for diversity and the desire to maintain coherence with human-like decision-making. 

\section{Discussion}

We compared the voting behaviors of LLMs and humans using voting data from real-world PB experimental study. 

Our results reveal distinct differences in human and LLM voting patterns and collective outcomes. Humans display significant variation in the number of approved projects, while LLM agents tend to approve fewer and exhibit more uniform behavior, even with the addition of varying personas. This insight is consistent with the finding of \citep{abdurahman2024perils} that synthetic AI-simulated sampling has a WEIRD (Western, Educated, Industrialized, Rich, Democratic) bias and often fail to show meaningful variance (or diversity) in their judgments.

The results also show that the choice of voting method influence LLM votes more so than it affects humans votes. Humans demonstrated proficiency in ranking and relative comparisons. For the LLM agents, GPT-4 excelled in quantifying and distributing preferences using cumulative voting,
whereas LLaMA-2 notably showed inconsistent voting, with ranked voting outcome being the least consistent outcome. This suggests that for LLM agents to make collective decisions, cumulative voting, where agents can assign precise points, might be a better choice than ranked voting.

Additionally, list ordering also influenced the LLM voting. Although these agents often provide convincing justifications for their votes in their responses (Appendix E), a quantifiable change in collective votes reveals an inconsistency in their preferences. This highlights the importance of looking into aggregated votes on top of individual LLM responses in understanding LLM behavior in the democratic context. 

Without adding a persona, LLMs display distinct preferences not shared by humans. For example, LLaMa-2 prefers kids-related projects, while GPT-4 shows a strong inclination towards transportation projects. The addition of personas partly offsets these innate preferences but introduces new challenges, leading LLMs to base their votes predominantly on persona-specific reported districts and urban interests. In contrast, human votes rely less on these factors, suggesting that their choices reflect not only self-interest but are also driven by other-regarding preferences, such as considering the community's interest.

Several strategies proposed in the literature to improve LLM behavior did not help in aligning them more closely with humans: 
Chain-of-thought (CoT) reasoning neither aligns LLM individual preferences nor collective outcomes more with humans, consistent with previously reported results \citep{chuang2024wisdom}. Furthermore, varying the temperature introduces a trade-off between diversity and accuracy in human alignment. At a temperature of 1, where GPT-4 shows the highest similarity to human voting, the preference range is overly concentrated. At higher temperatures, LLM agents match the human preference diversity range but produce more random and less accurate votes.

Some aspects of LLMs could support human voting patterns in PB settings. Specifically, unlike human voters who are often less cost-conscious in their selection of projects, LLMs show a higher preference for low-budget projects. LLMs are also relatively effective in quantifying their preferences, as shown in the consistency of cumulative voting compared to the outcomes with other voting methods. This suggests that LLMs could potentially assist humans in decision-making processes involving cost considerations or more accurately present human preferences in a quantifiable manner.


\subsection{Limitations:} (1) Context Specificity: Our findings concerning participatory budgeting and multi-winner elections may not be so generalizable, as it does not directly apply to the more common single-winner election formats (such as presidential elections). (2) Order Effects: We find that LLM votes are susceptible to order effects. While humans also exhibit primacy and recency effects \citep{van2011human}, our PB voting data did not include a treatment for measuring these order effects in humans. As a result, we could not quantify the difference between humans and LLMs. Future research is needed to address this gap.

\subsection{Future Directions:} (1) Persona Construction: Our results show that personas align LLM preferences more closely with human preferences in terms of topic preference. However, they also cause a divergence from human preferences regarding the balance between self-interest and other-regarding preferences. Future research should focus on creating persona descriptions that balance both aspects to better mirror human voting behavior. (2) Performance and Diversity: Varying temperature revealed a trade-off between model performance and diversity in LLM voting. Future research should investigate the extent to which improved alignment with human votes justifies the lack of diversity among decision-making agents. (3) Exploring Various Election Types: We tested a multi-winner election format in a PB setting. Testing additional election formats in different contexts would broaden the understanding of LLM applicability in democratic processes.

\section{Conclusion}
This study analyzed and compared the voting behaviors of LLM agents and human voters using data from real-world Participatory Budgeting experimental study. Our findings reveal significant differences between LLM and human voting behaviors in several key areas: the number of projects approved, consistency across different voting methods, qualitative preferences, and the degree of self-interest exhibited in voting patterns.
We explored various strategies to align LLM behaviors more closely with human behaviors, but each strategy presented notable trade-offs. Incorporating personas increased similarity in preferences for certain topics but also made LLMs more self-interested compared to humans. Furthermore, no temperature setting effectively maintained both the diversity and the human-like nature of preferences.
A notable advantage of LLMs in the context of PB is their cost awareness and their ability to rationalize decisions. While humans often lack cost-consciousness in participatory budgeting settings, LLMs demonstrate a greater consideration of costs. Additionally, LLMs using the Chain-of-Thought approach can provide reasoning before voting decisions, making them potentially useful for explaining collective decision-making outcomes, which can be difficult for voters with varying opinions to understand.

Overall, we empirically quantified some the biases and limitations of using LLMs in a democratic context. While LLM agents with personas can be useful in predicting voting patterns, our results indicate that the biases and lack of diversity in LLM decisions require critical evaluation before their deployment in democratic decision-making processes. Instead, we propose that LLMs be more effectively utilized within a human-in-the-loop framework. In this context, LLMs can function as support tools, assisting humans in addressing specific deficiencies in their decision-making processes, such as enhancing cost-consciousness, overcoming common human cognitive biases, summarization, explanation, or providing nuances and additional context for public policy and public sentiment.

\section*{Impact Statement}
This study critically examines the use of LLMs in democratic systems, particularly in the voting process. It reveals that the current capabilities of LLMs are inadequate for capturing the full spectrum of human perspectives, which can threaten the integrity of democratic outcomes.
The findings highlight the need for strict ethical guidelines to manage the integration of AI into democratic contexts. A human-centered approach is essential to ensure that AI deployment supports rather than compromises the collective intelligence derived from diverse human preferences in society.

\section*{Acknowledgments}
JY would like to express gratitude to the Swiss National Science Foundation (SNSF) for the financial support provided for the previously conducted PB voting experiment in \citet{yang2024}, which was part of the National Research Programme NRP77 on Digital Transformation (project no. 187249). JY also thanks Dominik Peters, Regula Hänggli Fricker, and Evangelos Pournaras for their intellectual contributions. MK, CIH, and DH acknowledge the support from the ``CoCi: Co-Evolving City Life'' project, funded by the European Research Council (ERC) under the EU’s Horizon 2020 research and innovation program (grant no. 833168). DD acknowledges the support from the Distributed Intelligence and Technology for Traffic and Mobility Management (DIT4TraM) project, funded by the EU’s Horizon 2020 Research and Innovation Programme (grant no. 953783).


{
\bibliography{aaai24}
}

\appendix
\onecolumn

\section{Additional Figures}

\begin{figure}[H] 
    \centering
    \includegraphics[width=0.49\linewidth]{figures/appr_hist.png}
    \includegraphics[width=0.49\linewidth]{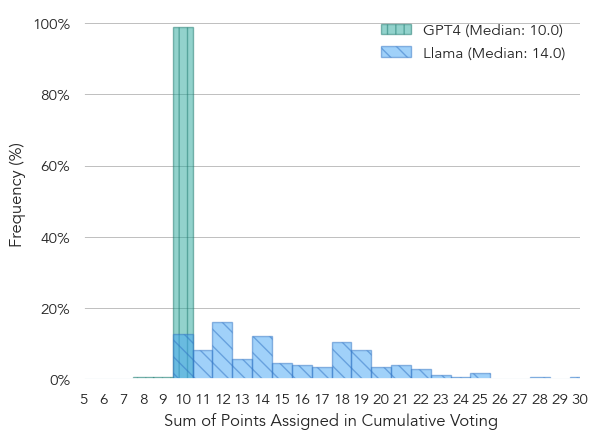}
    \caption{
    Histograms comparing the voting behavior of Human and LLMs. \textit{Left:} The number of selected projects in Approval Voting varies across the types of voters.
    \textit{Right:} Sum of assigned points in Cumulative Voting for GPT4 and LLaMA-2 voters. GPT4 adheres strictly to the instruction of assigning 10 points, LLaMA-2 does not.
    }
    \label{fig:appr_hist_cumulative_voting_distribution}
\end{figure}

\begin{figure}[H]
    \centering
    \includegraphics[width=0.6\linewidth]{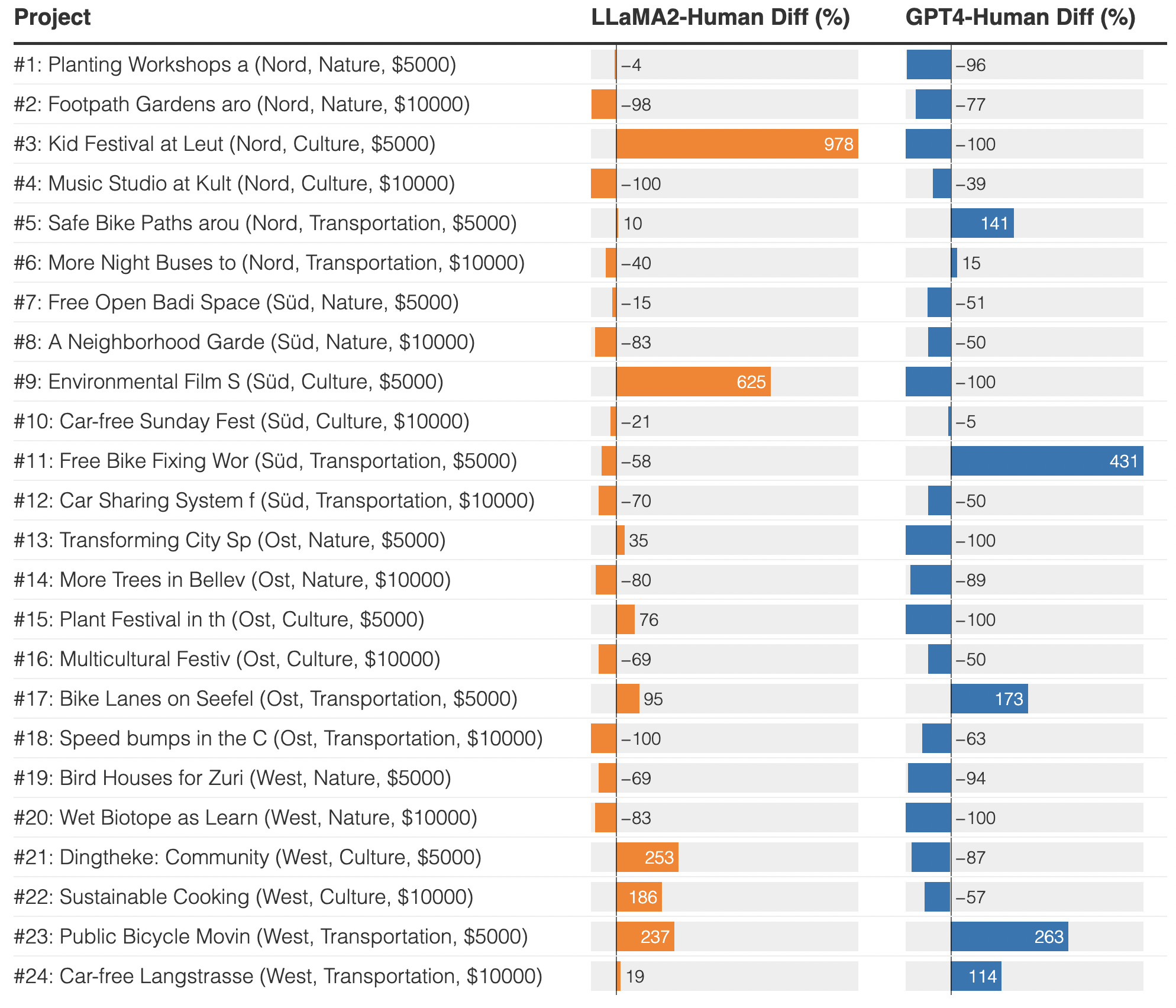}
    \caption{Comparative analysis of voting outcomes, highlighting the percentage differences between human and LLM voters. The 24 projects, spanning various districts and categories, are listed with corresponding vote percentage differences. A positive percentage indicates a project's greater priority for LLM voters compared to human voters.}
    \label{fig:human_diff_detailed}
\end{figure}

\begin{figure}[H]
    \centering
    \includegraphics[width=0.6\linewidth]{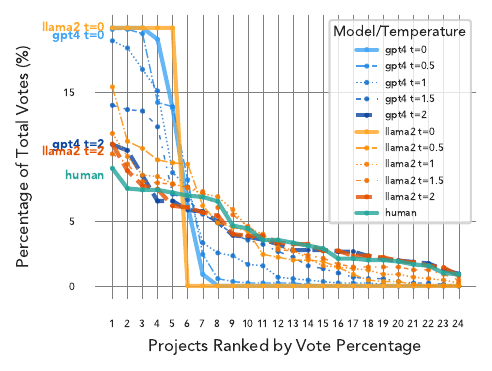}
    \caption{Vote percentage distribution across projects ranked by collective preference for human, GPT-4 and LLaMA-2 voters at varying temperature. The lines represent the normalized distribution of votes for each entity, highlighting the decision-making patterns and diversity of preferences.}
    \label{fig:temperature_preferences}
\end{figure}

\section{Additional Tables}

\begin{table}[H]
\centering
\caption{Pearson correlation between the ranking order and project IDs across various voting scenarios. In the LLM experiments, settings where project lists with reversed IDs (rev\_id) and reversed order (rev\_order) are included.}
\label{tab:vote_id_correlation}
\resizebox{0.4\columnwidth}{!}{%
\begin{tabular}{lrrr}\\
\toprule
 & Correlation & P-Value & Significant \\
 \midrule
human & 0.100 & 0.642 & False \\
gpt4 & 0.013 & 0.952 & False \\
gpt4\_rev\_id & -0.089 & 0.680 & False \\
gpt4\_rev\_order & -0.020 & 0.926 & False \\
llama2 & -0.092 & 0.668 & False \\
llama2\_rev\_id & 0.189 & 0.377 & False \\
llama2\_rev\_order & 0.331 & 0.114 & False \\
\bottomrule
\end{tabular}
}
\end{table}

\newpage
\section{Persona Description Construction}
\label{appendix:persona}

We construct persona descriptions from participant responses in a survey. Each entry consists of the participant's ID, urban category preferences, district information, connection to the city, and importance ratings.

\subsection*{Data Structure and Example Entry:}
\begin{table}[htbp]
\caption{Example Survey Response for Persona Description Construction}
\centering
\footnotesize
\begin{tabular}{l l}\\
\toprule
\textbf{Field} & \textbf{Value} \\
\midrule
ID & YK3TDKDG \\
1st Urban Category & Transport \\
1st Score & 60 \\
2nd Urban Category & Nature \\
2nd Score & 30 \\
3rd Urban Category & Culture \\
3rd Score & 10 \\
District & Nord \\
How Connected & A lot \\
District Importance & Moderately important \\
Urban Category Importance & Very important \\
Cost Importance & Not important at all \\
Likelihood Importance & Moderately important \\
\bottomrule
\end{tabular}
\end{table}

Based on this data, we then create a narrative description for each persona. The process of persona generation involves processing survey data and using the numerical scores to generate descriptive human language labels that reflect the participants' preference intensity. 

In the survey, participants rated the importance of each urban category, with the total score summing to exactly 100. To construct personas that reflect these nuances, the function interprets scores: if the importance score exceeds 50, it uses the descriptor ``having a strong preference for" the category. Scores between 33 and 50 lead to ``prefer,'' indicating a moderate preference. Scores below 33 result in no specific preference being highlighted. This nuanced interpretation ensures that each persona's narrative description accurately mirrors their interests and priorities in urban topics. 

The function further combines this with participants' district preferences and their prioritization of decision factors like district importance, urban category significance, and cost, creating a comprehensive and personalized narrative for each participant.

\subsection*{Corresponding Persona Description:}
\begin{quote}\small\sffamily
You are a university student from Nord district in Zurich. In urban 
topics, you have a strong preference for transport. When deciding on projects, 
you find the district to be moderately important, the urban category very 
important, and the cost of the project not important at all.
\end{quote}

\clearpage

\section{List of Projects}
\begin{table}[H]
\centering
\caption{List of 24 projects presented in the participatory budgeting experiment, as provided to both the 180 human participants and the LLM agents.}
\label{tab:project_list}
\footnotesize
\begin{tabular}{lllll}
\toprule
Id & Name & District & Category & Cost \\
\midrule
\#1 & Planting Workshops at Oerlikon & Nord & Nature & 5000\\
\#2 & Footpath Gardens around Train Stations & Nord & Nature & 10000\\
\#3 & Kid Festival at Leutschenpark & Nord & Culture & 5000\\
\#4 & Music Studio at Kulturbahnhof Affoltern & Nord & Culture & 10000\\
\#5 & Safe Bike Paths around Irchel Park & Nord & Transportation & 5000\\
\#6 & More Night Buses to Oerlikon & Nord & Transportation & 10000\\
\#7 & Free Open Badi Space in Wollishofen & Süd & Nature & 5000\\
\#8 & A Neighborhood Garden for Wiedikon & Süd & Nature & 10000\\
\#9 & Environmental Film Session for Kids & Süd & Culture & 5000\\
\#10 & Car-free Sunday Festivals on Mutschellenstrasse & Süd & Culture & 10000\\
\#11 & Free Bike Fixing Workshops & Süd & Transportation & 5000\\
\#12 & Car Sharing System for Young People & Süd & Transportation & 10000\\
\#13 & Transforming City Spaces under Trees into Gardens & Ost & Nature & 5000\\
\#14 & More Trees in Bellevue \& Sechseläutenplatz & Ost & Nature & 10000\\
\#15 & Plant Festival in the City Centre & Ost & Culture & 5000\\
\#16 & Multicultural Festival at Sechseläutenplatz & Ost & Culture & 10000\\
\#17 & Bike Lanes on Seefeldstrasse & Ost & Transportation & 5000\\
\#18 & Speed bumps in the City and the Lake Front & Ost & Transportation & 10000\\
\#19 & Bird Houses for Zurich-Höngg & West & Nature & 5000\\
\#20 & Wet Biotope as Learning Garden in Frankental & West & Nature & 10000\\
\#21 & Dingtheke: Community Things Exchange Library in Wipkingen & West & Culture & 5000\\
\#22 & Sustainable Cooking Workshop with Kids & West & Culture & 10000\\
\#23 & Public Bicycle Moving Trailer to be Borrowed & West & Transportation & 5000\\
\#24 & Car-free Langstrasse & West & Transportation & 10000\\
\bottomrule
\end{tabular}
\end{table}

\section{Example LLM Responses}
\label{appendix:llm-responses}

\subsection*{GPT-4 Responses: agent \#0, temp=1, no persona}
\begin{itemize}
    \item \textbf{Approval Voting:}
    \begin{lstlisting}[numbers=none]
    #5, #11, #17, #23
    \end{lstlisting}

    \item \textbf{5-Approval Voting:}
    \begin{lstlisting}[numbers=none]
    #2, #5, #11, #17, #24
    \end{lstlisting}

    \item \textbf{Cumulative Voting:}
    \begin{lstlisting}[numbers=none]
    #5 - 3 points #11 - 2 points #17 - 3 points #23 - 2 points 
    As a university student, I am focused on sustainable transportation and fostering a bike-friendly city, hence my preference for safe bike paths, free bike fixing workshops, bike lanes, and a public bicycle moving trailer. These projects would directly benefit my daily life and the lives of my peers, promoting health, sustainability, and convenience.
    \end{lstlisting}

    \item \textbf{Ranked Voting:}
    \begin{lstlisting}[numbers=none]
    #10,#22,#18,#24,#7  
    1. #10 Car-free Sunday Festivals on Mutschellenstrasse - Cultural events are important for community building and reducing traffic. 
    2. #22 Sustainable Cooking Workshop with Kids - Education on sustainability, especially for kids, is crucial for the future. 
    3. #18 Speed bumps in the City and the Lake Front - Safety is important, and speed bumps can help reduce accidents. 
    4. #24 Car-free Langstrasse - Reducing traffic can improve quality of life and it's an initiative towards a greener city. 
    5. #7 Free Open Badi Space in Wollishofen - Public spaces like this contribute to the well-being and are essential for the community during summer.
    \end{lstlisting}
\end{itemize}

\subsection*{LLaMA-2 Responses: agent \#3, temp=1, no persona}
\begin{itemize}
    \item \textbf{Approval Voting:}
    \begin{lstlisting}[numbers=none]
    Sure! As a university student in Zurich, I would like to vote for the following projects:  
    #5 Safe Bike Paths around Irchel Park - This project would improve the safety of bike paths in the area, making it easier for students like me to get around by bike. 
    #10 Car-free Sunday Festivals on Mutschellenstrasse - This project would create a fun and environmentally-friendly event that I could enjoy with my friends and fellow students. 
    #17 Bike Lanes on Seefeldstrasse - This project would improve the infrastructure for cyclists in the city, making it easier for me to get around by bike. 
    #22 Sustainable Cooking Workshop with Kids - This project would promote sustainability and healthy eating habits among young people, which is an important issue for me as a student.  
    I hope that my votes will help make these projects a reality!
    \end{lstlisting}

    \item \textbf{5-Approval Voting:}
    \begin{lstlisting}[numbers=none]
    Sure! Here are my votes:  
    #3 - Kid Festival at Leutschenpark (Nord) - I think it's important to invest in events that bring the community together and provide fun activities for kids.  
    #6 - More Night Buses to Oerlikon (Nord) - As a student, I often have late nights studying or socializing, and having more night buses would make it easier for me to get home safely.  
    #13 - Transforming City Spaces under Trees into Gardens (Ost) - I love the idea of turning unused spaces into beautiful gardens. It would make the city more enjoyable to live in and help the environment.  
    #17 - Bike Lanes on Seefeldstrasse (Ost) - As a cyclist, I think it's important to have designated bike lanes to ensure safety and encourage more people to cycle.  
    #22 - Sustainable Cooking Workshop with Kids (West) - I believe it's important to teach kids about sustainability and healthy eating habits from a young age. This project aligns with my values and interests.
    \end{lstlisting}

    \item \textbf{Cumulative Voting:}
    \begin{lstlisting}[numbers=none]
    Sure! Here are my preferences for the participatory budgeting program in Zurich, with three projects that I believe would greatly benefit the city and its residents:  
    1. #5 - Safe Bike Paths around Irchel Park (Nord) - 4 points As a university student, I believe that improving cycling infrastructure is crucial for a safer and more sustainable commute. This project would enhance the safety of bike paths around Irchel Park, making it easier for students and residents to get around the city without relying on cars. 
    2. #17 - Bike Lanes on Seefeldstrasse (Ost) - 3 points This project would create dedicated bike lanes on a busy street in the eastern part of Zurich, providing a safer and more convenient route for cyclists. As a student, I believe that investing in cycling infrastructure is essential for reducing traffic congestion and promoting sustainable transportation. 
    3. #22 - Sustainable Cooking Workshop with Kids (West) - 3 points This project would offer cooking workshops for children, focusing on sustainable food practices and healthy eating habits. As a young adult, I believe that it's important to educate the next generation about the impact of their food choices on the environment and their health. By supporting this project, I hope to encourage sustainable living practices among young people in Zurich.  
    Total points allocated: 10
    \end{lstlisting}

    \item \textbf{Ranked Voting:}
    \begin{lstlisting}[numbers=none]
    Sure! Here are my votes:  
    #1 - Planting Workshops at Oerlikon (Nord) - Nature - 5000 CHF 
    #2 - Free Open Badi Space in Wollishofen (S\u00fcd) - Nature - 5000 CHF 
    #3 - Environmental Film Session for Kids (S\u00fcd) - Culture - 5000 CHF 
    #4 - Car-free Sunday Festivals on Mutschellenstrasse (S\u00fcd) - Culture - 10,000 CHF 
    #5 - Bike Lanes on Seefeldstrasse (Ost) - Transportation - 5,000 CHF  
    My ranking is:  
    1. #1 - Planting Workshops at Oerlikon (Nord) - Nature - 5000 CHF 
    2. #3 - Environmental Film Session for Kids (S\u00fcd) - Culture - 5,000 CHF 
    3. #4 - Car-free Sunday Festivals on Mutschellenstrasse (S\u00fcd) - Culture - 10,000 CHF 
    4. #2 - Free Open Badi Space in Wollishofen (S\u00fcd) - Nature - 5,000 CHF 
    5. #5 - Bike Lanes on Seefeldstrasse (Ost) - Transportation - 5,000 CHF  
    I ranked projects based on my personal interest in environmental sustainability, cultural events, and transportation infrastructure. I believe that investing in planting workshops, environmental education for kids, and car-free festivals will contribute to a greener and more livable city. Additionally, I think it's important to provide free open spaces for residents to enjoy, especially in densely populated areas like Wollishofen. Finally, I support the development of bike lanes on Seefeldstrasse to promote sustainable transportation options.
    \end{lstlisting}
\end{itemize}

\subsection*{GPT-4 Responses: agent \#0, temp=1, with persona}
\begin{itemize}
    \item \textbf{5-Approval Voting:}
    \begin{lstlisting}[numbers=none]

    #19,#20,#21,#22,#7
    \end{lstlisting}
\end{itemize}

\subsection*{LLaMA-2 Responses: agent \#3, temp=1, with persona}
\begin{itemize}
    \item \textbf{5-Approval Voting:}
    \begin{lstlisting}[numbers=none]
    Sure! Based on my preferences, I would vote for the following projects:  
    #5 Safe Bike Paths around Irchel Park (Nord, Transportation, 5000 CHF) 
    #6 More Night Buses to Oerlikon (Nord, Transportation, 10000 CHF) 
    #17 Bike Lanes on Seefeldstrasse (Ost, Transportation, 5000 CHF) 
    #23 Public Bicycle Moving Trailer to be Borrowed (West, Transportation, 5000 CHF) 
    #24 Car-free Langstrasse (West, Transportation, 10000 CHF)  
    These projects align with my interests in transportation and improve the city's infrastructure for cyclists and public transportation.
    \end{lstlisting}
\end{itemize}

\subsection*{GPT-4 Chain-of-Thought Responses: agent \#1, temp=1, with persona}
\label{sec:cot-ex}
\begin{itemize}
    \item \textbf{First thought response}
    \begin{lstlisting}[numbers=none] 
    As a university student from the Nord district in Zurich with a strong preference for transport within urban topics, my decisions in a participatory budgeting program would be heavily influenced by proposals tackling issues of mobility and connectivity. Hence, I would favor projects that aim to improve the transportation infrastructure, such as the development of bike lanes, enhancement of public transport services, and the integration of sustainable transport solutions. Projects that specifically target improvements within the Nord district would be of slightly less importance to me than the overall category of urban transport, but I would still consider them favorably, especially if they promise to better the daily commute and accessibility for residents, including my fellow students. The cost of the project would not influence my decision, as I am focused on the potential benefits and improvements in the transport sector rather than financial constraints. My main criterion for voting for a project would be its potential to make travel within Zurich more efficient, safer, and environmentally friendly.
    \end{lstlisting}

    \item \textbf{5-Approval Voting}
    \begin{lstlisting}
    Considering my strong preference for transport projects, I would allocate my votes to the following:  
    #5, Safe Bike Paths around Irchel Park 
    #6, More Night Buses to Oerlikon 
    #11, Free Bike Fixing Workshops 
    #17, Bike Lanes on Seefeldstrasse 
    #24, Car-free Langstrasse  
    These projects support an improvement in transportation, which aligns closely with my personal interests. Four of these projects are in my home district (Nord), further reinforcing my moderate preference for district-related improvements. Even though project #24 is not in my district, it supports the transport category, which I consider very important.
    \end{lstlisting}
\end{itemize}

\subsection*{GPT-4 Chain-of-Thought Responses: agent \#39, temp=1, with persona}

\begin{itemize}
    \item \textbf{First thought response}
    \begin{lstlisting}[numbers=none] 
    As a university student from the West district of Zurich with a strong preference for culture, my decisions regarding urban projects in a participatory budgeting program would be influenced by how well the projects align with the cultural enrichment of my district and the city at large. Given my background, I would prioritize funding for projects that promote cultural activities, such as the establishment of art galleries, support for local theaters, music festivals, or public art installations, especially those that are within or benefit the West district, as I find the district of execution very important.Moreover, since the urban category is very important to me, I would look favorably on projects that enhance urban livability through cultural means, such as creating cultural community centers or artist-in-residence programs that could bring a diverse range of cultural experiences to the residents.The cost of the project is moderately important, suggesting that while I am cost-conscious and desire effective use of funds, I might be willing to support slightly more expensive projects if they offer significant cultural benefits to the community. Nevertheless, overly expensive projects might be deprioritized in favor of those that offer better cost-to-benefit ratios within my interests. To sum up, cultural relevance, location within the West district, and a reasonable cost structure would guide my decisions in the participatory budgeting process.
    \end{lstlisting}

    \item \textbf{5-Approval Voting}
    \begin{lstlisting}[numbers=none] 
    Based on my preferences for cultural projects, their location in the West district, and moderate consideration of the project cost, I would select the following:
    #19 - Bird Houses for Zurich-H\u00f6ngg (West, Nature, 5000)
    #20 - Wet Biotope as Learning Garden in Frankental (West, Nature, 10000)
    #21 - Dingtheke: Community Things Exchange Library in Wipkingen (West, Culture, 5000)
    #22 - Sustainable Cooking Workshop with Kids (West, Culture, 10000)
    #24 - Car-free Langstrasse (West, Transportation, 10000)
    These projects reflect a strong emphasis on cultural enrichment and environmental considerations in my home district.
    \end{lstlisting}
\end{itemize}

\end{document}